\documentclass[conference]{IEEEtran}
\ifCLASSINFOpdf
\else
\fi
\usepackage{blindtext}
\usepackage[T1]{fontenc}
\usepackage[utf8]{inputenc}
\usepackage{graphicx} 
\usepackage{amssymb} 
\usepackage{amsmath} 
\usepackage{makecell} 
\usepackage{siunitx} 
\usepackage{subcaption} 

\usepackage{pdfpages} 
\usepackage{hyperref} 
\usepackage{rotating}
\usepackage{multirow}
\usepackage{cite}

\usepackage[absolute]{textpos}
\usepackage{everyshi}

\IEEEoverridecommandlockouts 

\usepackage[font=scriptsize]{caption}

\begin{document}


\title{CNN-based Approaches For\\Cross-Subject Classification in Motor Imagery:\\From The State-of-The-Art to DynamicNet}

\author{Alberto Zancanaro$^{1}$, Giulia Cisotto$^{1,2,3}$,  João Ruivo Paulo$^{4}$, Gabriel Pires$^{4,5}$, and Urbano J. Nunes$^{4,6}$ \\
\\
$^1$Dept. Information Engineering, University of Padova, Italy\\
$^2$National Centre for Neurology and Psychiatry, Tokyo, Japan \\
$^3$National Inter-University Consortium for Telecommunications (CNIT), Padova, Italy\\
$^4$Institute of Systems and Robotics, University of Coimbra, Coimbra, Portugal\\
$^5$Engineering Department, Polytechnic Institute of Tomar, Tomar, Portugal\\
$^6$Department of Electrical and Computer Engineering, University of Coimbra, Coimbra, Portugal\\
alberto.zancanaro.1@studenti.unipd.it, giulia.cisotto.1@unipd.it, \{jpaulo, gpires, urbano\}@isr.uc.pt
}


%


\maketitle

\begin{abstract}
Motor imagery (MI)-based brain-computer interface (BCI) systems are being increasingly employed to provide alternative means of communication and control for people suffering from neuro-motor impairments, with a special effort to bring these systems out of the controlled lab environments.
Hence, accurately classifying MI from brain signals, e.g., from electroencephalography (EEG), is essential to obtain reliable BCI systems. However, MI classification is still a challenging task, because the signals are characterized by poor signal-to-noise-ratio, high intra-subject and cross-subject variability.
Deep learning approaches have started to emerge as valid alternatives to standard machine learning techniques, e.g., filter bank common spatial pattern (FBCSP), to extract subject-independent features and to increase the cross-subject classification performance of MI BCI systems.
In this paper, we first present a review of the most recent studies using deep learning for MI classification, with particular attention to their cross-subject performance. Second, we propose DynamicNet, a Python-based tool for quick and flexible implementations of deep learning models based on convolutional neural networks. We show-case the potentiality of DynamicNet by implementing EEGNet, a well-established architecture for effective EEG classification. Finally, we compare its performance with FBCSP in a 4-class MI classification over public datasets. To explore its cross-subject classification ability, we applied three different cross-validation schemes.
From our results, we demonstrate that DynamicNet-implemented EEGNet outperforms FBCSP by about 25\%, with a statistically significant difference when cross-subject validation schemes are applied. 
We conclude that deep learning approaches might be particularly helpful to provide higher cross-subject classification performance in multi-class MI classification problems with respect to FBCSP. Moreover, in the future DynamicNet could be useful to implement new architectures to further investigate cross-subject classification for MI BCI in real-world scenarios.
\end{abstract}

\begin{textblock*}{17cm}(1.7cm, 0.5cm)
\noindent\scriptsize This work has been submitted to the IEEE for possible publication. Copyright may be transferred without notice, after which this version may no longer be accessible.\\
\textbf{Copyright Notice}: \textcopyright 2021 IEEE. Personal use of this material is permitted. Permission from IEEE must be obtained for all other uses, in any current or future media, including reprinting/republishing this material for advertising or promotional purposes, creating new collective works, for resale or redistribution to servers or lists, or reuse of any copyrighted component of this work in other works.
\end{textblock*}


%
\IEEEpeerreviewmaketitle


\section{Introduction}
\label{sec:1_Introduction}

Brain–computer interfaces (BCI) provide a direct communication channel between the brain and external devices~\cite{wolpaw_BCI,Silvoni2013}. One of the most common BCIs is based on motor imagery (MI), i.e. the imagination of movements (e.g., of hands, feet, tongue). MI-related EEG  has been extensively studied in the context of communication~\cite{BCI_comunication}, robotic control~\cite{BCI_wheelchair}, gaming~\cite{BCI_gaming} and  rehabilitation~\cite{BCI_rehabilitation}.
Despite of extensive research, MI-based BCIs still face many challenges. EEG signals have a low signal-to-noise ratio (SNR), can be easily corrupted by various artifacts (like eye movement or other muscular activity), have a poor spatial resolution leading to channel correlation, and the features extracted from the signals are very subject-dependent. This last point remains one of the biggest challenges in the  BCI field, making it almost mandatory to collect data and calibrate the BCI for each subject. The search for approaches that allow an efficient cross-subject capability is therefore an attractive area of research in the BCI field. Traditional machine learning approaches for MI classification have been mainly based on common spatial patterns (CSP)~\cite{CSP_original_paper} and its variants, such as the filter bank CSP (FBCSP)~\cite{FBCSP}, combined with classifiers such as support vector machines (SVM) or linear discriminant analysis (LDA). These approaches have been quite successful in classifying MI data for the within-subject case~\cite{FBCSP, FBCSP_results_citation}, but have poor performance in the cross-subject scenario~\cite{EEGNet_paper}. Due to this fact, recently, researchers started to study deep learning (DL) techniques to analyze and classify EEG data~\cite{EEGNet_paper, Schirrmeister_EEG_CNN, Sakhavi_EEG_CNN, Bressan2021}. DL research is also motivated by the success of these techniques in other fields, e.g., image classification, speech recognition and text analysis. Compared with traditional machine learning, DL offers a unique end-to-end solution that can automatically extract features that best classify raw EEG data.
However, the design of DL-based architectures can be cumbersome. The need to select and combine different layers in a network, along with the training and validation of the best design is still time-consuming and requires some expertise. This emphasizes the importance of providing flexible tools that allow a systematic implementation of different DL architectures to easily achieve classification models.
%
%
The main contributions of this work are: (1) to survey the most recent works that used DL to classify MI data, with particular attention to cross-subject classification; (2) to introduce \emph{DynamicNet}, a new open-source Python-based toolbox to implement a variety of DL models. Also, we show-case the potentiality of \emph{DynamicNet} by implementing EEGNet, a well-established DL architecture for effective EEG classification; (3) we compare its performance with FBCSP in a 4-class MI classification over a public dataset. To explore its cross-subject classification ability, we apply three different cross-validation schemes.


The structure of the paper is as follows. Section~\ref{sec:2_Survey} presents a survey of the state-of-the-art on cross-subject deep learning-based motor-related classification. Section~\ref{sec:3_Methodology} introduces \emph{DynamicNet} and its use to implement EEGNet. Section~\ref{sec:4_Results and discussions} presents and discusses the results. Finally, Section~\ref{sec:5_Conclusions} concludes the work.

\section{Cross-subject Deep Learning studies}
\label{sec:2_Survey}
We have conducted a survey over the most recent papers, within the last 5 years, i.e., 2017-2021 (no relevant peer-reviewed papers could be actually found in 2021), dealing with DL-based MI EEG classification.
Table~\ref{tab:paper_comparison} summarizes the main works we found and sorts them in a chronological order. For each work we report the DL-based classification model, the datasets, including their size and the number of subjects, the number of classes used in the cross-subject investigation (if stated), the preprocessing procedure, the training and validation strategy used for cross-subject investigation, and the performance achieved in terms of accuracy. We note that, depending on the number of classes used in the classification (2 or 4), the chance level accuracy differs from one study to the other. Also, it is worth mentioning that only two works~\cite{EEGNet_paper,Amin_MCNN} investigated cross-subject classification in the 4-class scenario. The remaining ones limited the dataset to a 2-class classification problem and accordingly reported results.

\begin{table*}
    \centering
    \caption{
        Most recent studies (2017-2021) using deep learning to classify MI data with attention to cross-subject classification. The symbol \textasciitilde{} indicates that the numerical value was extracted from a figure. The symbol "-" in the \textit{Number of classes used for cross-subject validation} row,  means that that paper perform only intra-subject classification. Regarding the results for intra-subject classification they were obtained using all the classes of the various datasets. The only exception is with the work of Xu et al. where they used 2 classes in both intra-subject and cross-subject classification.
        D2a = Dataset 2a BCI Competition IV, D2b = Dataset 2b BCI Competition IV, HGD =  High Gamma Dataset, BP = Bandpass filter, EMA = exponential moving average. 
        \\\hspace{\textwidth} $^1$ The peer-reviewed paper was published in 2018; however, a previous version was available in ArXiv since 2016. The authors also implemented FBCSP to compare their results with EEGNet, so the FBCSP results are also reported.
    }
    \label{tab:paper_comparison}
    \resizebox{\linewidth}{!}{%
    \begin{tabular}{
        |>{\centering\hspace{0pt}}m{0.11\linewidth}
        |>{\centering\hspace{0pt}}m{0.16\linewidth}
        |>{\centering\hspace{0pt}}m{0.16\linewidth}
        |>{\centering\hspace{0pt}}m{0.16\linewidth}
        |>{\centering\hspace{0pt}}m{0.16\linewidth}
        |>{\centering\hspace{0pt}}m{0.16\linewidth}
        |>{\centering\hspace{0pt}}m{0.16\linewidth}
        |>{\centering\arraybackslash\hspace{0pt}}m{0.125\linewidth}|} 
    \cline{2-8}
    \multicolumn{1}{>{\centering\hspace{0pt}}m{0.063\linewidth}|}
    {~} & 
    Schirrmeister et al., 2017~\cite{Schirrmeister_EEG_CNN}  & 
    Lawhern et al.,2018~\cite{EEGNet_paper}$^{1}$ & 
    Amin et at., 2019~\cite{Amin_MCNN} & 
    Borra et al., 2020~\cite{Borra_SincShallowNet} & 
    Xu et al., 2020~\cite{Xu_cross_dataset} & 
    Roots et al.,2020~\cite{EEGNet_Fusion}  & 
    An et al., 2020~\cite{An_FewshotNetwork}                                                \\ 
    \hline
    Cross Subject & 
    No  & 
    Yes & 
    Yes & 
    No  & 
    Yes & 
    Yes & 
    Yes 
    \\ 
    \hline
    Architecture(s)
      used                                           & ShallowNet \par{}
        Deep ConvNet \par{}
        (both architectures originally proposed in this paper)                                                                         & EEGNet \par{}
      (architecture originally proposed in this paper)                                                              & MCNN
      Fusion Net \par{}
        CCNN Fusion Net \par{}
        (Both architectures originally proposed in this paper)               & Sinc-ShallowNet
        (ShallowNet derivation)                                                                                                                                                & EEGNet \par{}
        ShallowNet                                                                                                                                                                                                                                                           & Fusion
      EEGnet \par{}
        (EEGNet derivation)                                                                                                                                                                   & RelationNet-Attention \par{}
      (architecture originally proposed in this paper)                                                                                                                                                                                \\ 
    \hline
    Dataset\par{}(\# subjects) & 
    D2a (9 subjects)\par{} HGD (20 subjects) & 
    Various (included D2a) & 
    D2a (9 subjects)\par{} HGD (20 subjects)& 
    D2a (9 subjects)\par{} HGD (20 subjects)& 
    Various (included D2a, D2b and eegmmidb) & 
    eegmmidb (103 subjects) & 
    D2a (9 subjects, only for testing)\par{}D2b (8 subjects, Training and testing)  
    \\ 
    \hline
    Number of classes used for intra-subject validation & 
    4 & 
    4 & 
    4 & 
    4 & 
    2 & 
    - & 
    -                                                                             \\ 
    \hline
    Number of classes used for cross-subject validation & 
    - & 
    4 & 
    4 & 
    - & 
    2 & 
    2 & 
    2                                                                             \\ 
    \hline
    Pre-processing or input representation                           & BP
      4-38Hz with 3rd order Butterworth (D2a);
        BP 4-125Hz with third order Butterworth (HGD);
        Standardized with EMA with a decay factor of 0.999 (both) & Resampled
      at 128Hz;
        BP 4-40Hz with 3rd order Butterworth;
        Standardized with EMA with a decay factor of 0.999 & -                                                                                                             & Downsampled
      from 500Hz to 250Hz, BP 4-125Hz with 3rd order Butterworth standardized with
      EMA with a decay factor of 0.999 (HGD);
        BP 4-38Hz with 3rd order Butterworth (D2a);
        ~ & Pre-Aligment
      Strategy based on Riemannian geometry                                                                                                                                                                                                                            & BP
      2-60Hz of 5th order;
        Notch filter at 60Hz                                                                                                                                                        & -                                                                                                                                                                                                                                                       \\ 
    \hline
    Cross subject
    training and validation strategy & 
    - & 
    Training set: 5 subjects;\par{} 
        Validation set: 3 subjects;\par{} 
        Test set: 1 subject & 
    Leave one subject out.\par{}
        Training set: 8 subject;\par{}
        Test set: 1 subject. (the one not used for training) & 
    - & 
    Leave one subject out.\par{}
        Training set: 8 subject;\par{}
        Test set: 1 subject (the one not used for training). \par{}
        Since they used mutiple dataset to make it possible to compare the various results, only 3 channels were chosen for each dataset. & 
    Training set: 70\% of the data; \par{}
        Validation set: 10\% of the data; \par{}
        Test set: 20\% of the data. \par{}
        The data from different subjects were mixed altogether before the training-test split. &
    Leave one subject out.\par{}
        Training set: 8 subject; \par{}
        Test set: 1 subject. \par{}
        For the D2a only 3 channels were used (C3, CZ, C4). The results regarding D2a were obtained with the model trained on data of D2b.  \\ 
    \hline
    
    Average intra subject accuracy & 
    71.9\% (D2a - ShallowNet) \par{}70.1\% (D2a - Deep ConvNet) \par{}93.9 (HGD - ShallowNet) \par{} 91.4\% (HGD - DeepConv Net) \par{} 91\% (HGD - FBCSP) & 
    \textasciitilde{}67\% (EEGNet) \par{} \textasciitilde{}67\% (FBCSP) & 
    75.7\% (MCNN - D2a) \par{}73.8\% (CCNN - D2a) \par{} 95.4\% (MCNN - HGD) \par{} 93.2\% (CCNN - HGD) & 
    72.8±12.9\% (D2a) \par{}91.2±9.1\% (HGD)  & 
    79\% (D2a - EEGNet) \par{}77\%(D2a - ShallowNet) \par{} 80\% (D2b - EEGNet) \par{}79\%(D2b - ShallowNet) \par{}56\% (eegmmidb - EEGNet)\par{}56\%(eegmmidb - ShallowNet) & 
    -  & 
    -  
    \\ 
    \hline
    Average cross subject accuracy~ & 
    - & 
    \textasciitilde{}40\% (EEGNet) \par{} \textasciitilde{}32\% (FBCSP) &
    42.1\% (MCNN - D2a) \par{}55.3\% (CCNN - D2a) \par{}71.4\% (MCNN - HGD) \par{}79.2\% (CCNN - HGD) & 
    - & 
    \textasciitilde{}70\% (D2a - EEGNet) \par{}\textasciitilde{}70\%(D2a - ShallowNet) \par{} \textasciitilde{}76\% (D2b - EEGNet) \par{}\textasciitilde{}75\%(D2b - ShallowNet) \par{}\textasciitilde{}69\% (eegmmidb - EEGNet) \par{}\textasciitilde{}69\%(eegmmidb - ShallowNet) & 84.1\% (ME task) \par{} 83.8\% (MI~ task) & 
    74.6±10.18\% (D2b) \par{}59.1±11.1\% (D2a)                               
    \\
    \hline
    \end{tabular}
    }
\end{table*}

\subsection{Deep Learning Models}
From our survey, we found that most articles employ convolutional neural network (CNN)-based DL models. Such models generally implement an horizontal convolution, followed by a vertical convolution. The models act as the cascade of a frequency filtering stage, followed by a spatial filtering stage, i.e., a processing very similar to the the gold-standard FBCSP.

One of the most famous DL models for EEG data is the \emph{EEGNet} proposed by Lawhern \emph{et al.}~\cite{EEGNet_paper} in 2018. EEGNet is a CNN inspired by the FBCSP approach that uses horizontal and vertical convolution to simulate frequency and spatial filtering (see Section~\ref{subsec:3_EEGNet} for more details). Its original version, as well as several variants, have been recently applied to MI data.
Among other \emph{EEGNet} implementations, the \emph{Fusion-EEGNet}~\cite{EEGNet_Fusion} and the \emph{Incep-EEGNet}~\cite{incpet_EEGNet_paper} gained much attention and have been employed for different classification purposes, as shown in Table~\ref{tab:paper_comparison}. %
\emph{ShallowNet}~\cite{Schirrmeister_EEG_CNN}  also has a similar architecture, with the main difference being the use of another activation function for the last layers and the lack of depthwise convolutions.

Interestingly, Amin et at.~\cite{Amin_MCNN} proposed a multilevel CNN-based architecture with feature fusion at different layers of the CNN architecture. Their model could robustly learn time and frequency-domain features thanks to the application of horizontal and vertical convolutions, respectively. Then, exploiting the fusion of different abstract representations (i.e., from different CNN layers), it could outperform the state-of-art. Besides, fusion was alternatively obtained via feed-forward network (namely, in the MCNN model) or via autoencoder (namely, in the CCNN model).
%

\subsection{Datasets and Preprocessing}
The most commonly used datasets for motor-related EEG classification are the \emph{dataset 2a} proposed at the BCI Competition IV~\cite{dataset_BCI_competition} (see section~\ref{subsec:dataset} for more details), and the \emph{High Gamma Dataset} (HGD)~\cite{Schirrmeister_EEG_CNN}.
While the former includes MI data from 9 healthy subjects, the latter is a collection of motor execution (ME) EEG data from 20 healthy individuals. 
The wide use of these two datasets in the BCI community allows us to provide a fair comparison among the different DL models used to classify them.
Additionally, the \emph{dataset 2b} from the BCI competition IV (D2b, 2 classes~\cite{dataset_BCI_competition}), and the \emph{PhysionNet eegmmidb} (109 subjects, 4 classes, both MI and ME task~\cite{eegmidb}) were used in a few works. 

Finally, similar preprocessing has been applied across the surveyed papers: band-pass filtering (with variable choice of the cut-off frequencies) and normalization (e.g., the exponential moving~\cite{Schirrmeister_EEG_CNN}) were mostly used.

\subsection{Classification Accuracy}
Table~\ref{tab:paper_comparison} reports the results for both intra-subject and cross-subject classification.

Intra-subject classification performance was mostly obtained from the dataset 2a with 4 classes. In this scenario, FBCSP reaches an accuracy of 67\% \cite{EEGNet_paper}, while DL models achieve 67\% with EEGNet~\cite{EEGNet_paper} up to 75.7\% with MCNN~\cite{Amin_MCNN}.
In the 4-class HGD, FBCSP reaches an accuracy around 91\% \cite{Schirrmeister_EEG_CNN}, while DL models vary between 91.2\% with Sinc-ShallowNet~\cite{Borra_SincShallowNet} up to 95.4\% with MCNN~\cite{Amin_MCNN}.
Thus, we can conclude that DL models outperform FBCSP in the 4-class problem.
It is worth to note that better results have been obtained for HGD because it contains ME data, making the classification easier compared to MI data. 


On the other hand, in the cross-subject classification, the results are strongly influenced by the number of classes (2 or 4 classes) and the cross-subject training and validation strategy.
%
For the 4-class problem, Lawhern et al.~\cite{EEGNet_paper} obtained a cross-subject accuracy around 40\% using the dataset 2a. Despite such low accuracy, it still outperformed FBCSP (accuracy around 32\%). With the same dataset, Amin et al.~\cite{Amin_MCNN} reached a better accuracy (55.3\%) with the CCNN model.
With the HGD, including ME data, the cross-subject classification increased up to 79.2\% (obtained with the CCNN model).
%
When only 2 classes are considered, we observe a general increase in the accuracy values, as expected, given the simplification of the classification task.
Xu et al.~\cite{Xu_cross_dataset} presented an extensive study on cross-subject with 2 different architectures (ShallowNet and the EEGNet) over 8 datasets.

Overall, despite the relatively good results of the binary cross-subject classification, the accuracy values in the 4-class remain generally low, with ample room for improvement.

Interestingly, to the best of our knowledge, \cite{Xu_cross_dataset} and~\cite{An_FewshotNetwork} are the only works that provide results for cross-dataset classification (i.e., the models were trained on one dataset and tested on another one).
For example, in~\cite{An_FewshotNetwork}, the authors trained a model on D2b and tested it on the data of D2a.

\subsection{Cross-Subject Training Strategy}
We investigated cross-subject classification by analyzing the cross-validation strategy used in the surveyed papers.
We can observe that the main cross-validation strategy was the \textit{"leave-one-subject-out"}: given a dataset with $n$ subjects, then $n-1$ subjects were used as the training set, while the remaining one represented the test set.
Alternatively, a training-validation-test strategy was found: in this case, one subject was kept for the test, while the $n-1$ subjects were split into the training set and the validation set. This was the scheme used by Lawhern et al.~\cite{EEGNet_paper}, who considered 5 subjects to train their model, 3 of them for the validation, and 1 subject for testing (they used the dataset 2a, including 9 subjects). Differently, Roots et al.~\cite{EEGNet_Fusion} mixed the data from all subjects together (this approach is rarely seen in the literature) and a 70\%-20\%-10\% split was implemented for training, validation and test sets, respectively.
%

\section{Motor Imagery Classification}
\label{sec:3_Methodology}

\subsection{DynamicNet}
\begin{figure}
    \centering
    \includegraphics[width = \linewidth]{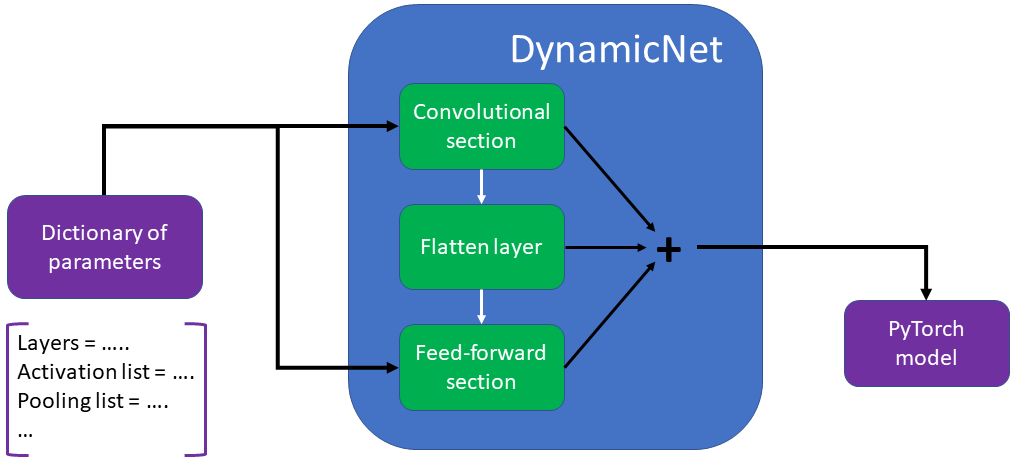}
    \caption{Diagram overview of DynamicNet. The tool generally constructs the model in 3 step. (1) iterates through the parameters of CNN section and builds it. (2) automatically evaluates flatten dimension and builds the flatten layer. (3) like (1) but for the feed-forward section. In case you want only a CNN (no flatten layer) or only a feed-forward network then only the relative step will be performed.}
    \label{fig:DynamicNet general scheme}
\end{figure}

\emph{DynamicNet} is a novel open-source toolbox based on PyTorch that allows to quickly implement a variety of DL models.
%
%
As sketched in Fig.~\ref{fig:DynamicNet general scheme}, \emph{DynamicNet}\footnote{GitHub repository DynamicNet: \href{https://github.com/jesus-333/Dynamic-PyTorch-Net}{https://github.com/jesus-333/Dynamic-PyTorch-Net}} allows to create a model by simply defining a Python dictionary containing general parameters of a CNN architecture, e.g., the number of convolutional layers, the list of activation functions, and the list of pooling layers. The dictionary of parameters is given as an input to the class.
\emph{DynamicNet} allows a rapid construction and debugging of the DL model with the design of on-the-fly configurations for entire layers.

The design of a DL model can be divided into three phases: 
\begin{enumerate}
    \item The design of the convolutional section. In this phase, the \emph{DynamicNet} iterates through the parameters of the convolutional section and, at each iteration, a convolutional layer is built. Each convolutional layer has the same structure and consists of the sequence of a convolution, a normalization, an activation, a pooling and a dropout. For each of them, the parameters are included in the corresponding list (e.g., \textit{activation\_list}, \textit{dropout\_list} etc). At the $i$-th iteration, \emph{DynamicNet} takes the $i$-th parameters from each list and uses them to build the $i$-th layer.
    \item The design of the flatten layer. In this phase, \emph{DynamicNet} builds automatically the flatten layer to connect to the first layer of the feed-forward section. The number of neurons in the flatten layer is automatically defined during the design of the convolutional layer (previous step). The user must only specify the dimension of the input. This information is used to create a "dummy input" that is propagated through the convolutional layers. The output of the convolutional layers is then flattened and its dimension is used as the number of input neurons. 
    \item The design of the feed-forward section. This process is similar to the first phase, with the main difference that the parameters are related to the feed-forward section. In the cases where a simple feed-forward architecture is designed (\emph{DynamicNet} allows this simpler case), phases 1 and 2 are skipped.
\end{enumerate}

The current version of \emph{DynamicNet} can only be used to create feed-forward neural networks-based and CNN-based DL models. Future developments will possibly provide more complex modules (e.g., incept module), as well as the integration of explainability tools (e.g., gradCAM~\cite{gradCAM_paper} and SHAP~\cite{SHAP_paper}).

\subsection{Implementation of EEGnet using DynamicNet}
\label{subsec:3_EEGNet}

\begin{figure}
    \centering
    \includegraphics[width = \linewidth]{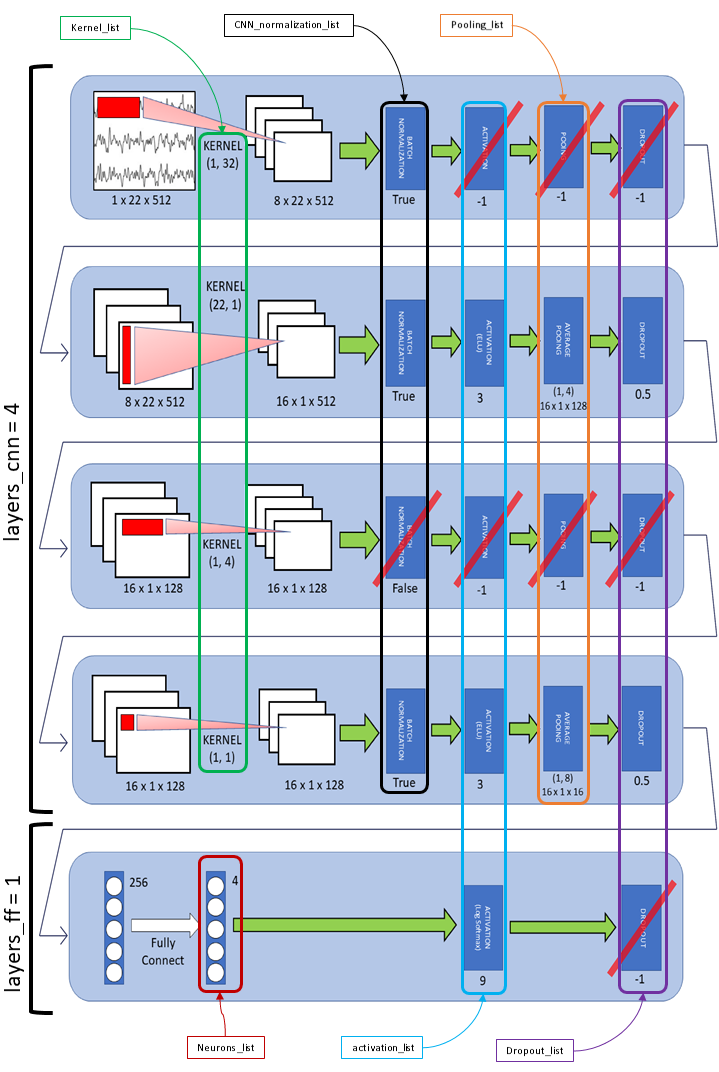}
    \caption{\emph{EEGNet} implementation with \emph{DynamicNet}.}
    \label{fig:EEGNet_DynamicNet}
\end{figure}

An example of how \emph{DynamicNet} can be used to implement \emph{EEGNet} is shown in Fig.~\ref{fig:EEGNet_DynamicNet}.
As mentioned above, the tool works cycling through all parameters and builds a layer at a time.  When all the layers are built, they are stacked all together. The complete list of parameters for \emph{EEGNet} can be seen in Table~\ref{tab:Dynamic Net_EEGNet implementation}. The convolutional section consists of 4 layers (\textit{layers\_cnn = 4}). Also, as it can be seen from the \textit{kernel\_list}, the first kernel performs an horizontal convolution (1,32) and the second one implements a vertical convolution (22,1). This step is then followed by another horizontal convolution  (1,4) and a point-wise convolution (1,1). The \textit{filters\_list} and the \textit{group\_list} specify how many input and output channels are in each layer, and how many filters are used in each layer. For example, the (1,8) in the \textit{filters\_list} means that an input with a single channel is fed to the model and is convoluted in 8 different channels, such that the output of that layer results in a depth of 8. Depth-wise convolution is achieved by setting the group of each layers equal to the number of input channels. As in~\cite{EEGNet_paper}, the activation, the pooling and the dropout were used only after the spatial convolution (layer 2) and the point-wise convolution (layer 4). The feed-forward section consists of 4 neurons (i.e., the desired number of classes) that are connected to the flattened output of the convolutional section.

\begin{table}
    \centering
    \caption{EEG building parameters using DynamicNet. A description of the meaning of each parameter can be found in the GitHub repository. The tuple in \textit{kernel\_list}, \textit{padding\_list} and \textit{pooling\_list} indicates the size of the kernel used in that operation and following the PyTorch convection are defined as \textit{height x width}. The tuple \textit{filter\_list} indicates the input and output channels (i.e. filters) for those layers.}
    \label{tab:Dynamic Net_EEGNet implementation}
    \resizebox{0.97\linewidth}{!}{%
    \begin{tabular}{|c|c|c|} 
        \cline{2-3}
        \multicolumn{1}{c|}{}                                                               & Dictionary key           & Value                                       \\ 
        \hline
        \multirow{10}{*}{\begin{tabular}[c]{@{}c@{}}Convolutionatl \\ section\end{tabular}} & h                        & 22                                          \\
                                                                                            & w                        & 512                                         \\
                                                                                            & layers\_cnn              & 4                                           \\
                                                                                            & kernel\_list             & {[}(1, 32), (22, 1), (1, 4), (1, 1)]         \\
                                                                                            & filters\_list            & {[}(1, 8), (8, 16), (16, 16), (16, 16)]     \\
                                                                                            & stride\_ist              & -                                           \\
                                                                                            & padding\_list            & {[}(0, 4), [0, 0], (0, 2), [0, 0]]          \\
                                                                                            & pooling\_list            & {[}-1, [1, (1, 4)], -1, [1, (1, 8)]]        \\
                                                                                            & groups\_list             & {[}1, 8, 16, 1]                             \\
                                                                                            & CNN\_normalization\_list & {[}True, True, False, True]                 \\ 
        \hline
        \multirow{2}{*}{\begin{tabular}[c]{@{}c@{}}Feed-Forward\\section\end{tabular}}      & layers\_ff               & 1                                           \\
                                                                                            & neurons\_list            & {[}4]                                       \\ 
        \hline
        \multirow{3}{*}{\begin{tabular}[c]{@{}c@{}}Common \\section\end{tabular}}                                                       & activation\_list         & {[}-1, 3, -1, 3, 9]                      \\
                                                                                            & bias\_list               & {[}False, False, False, False, True, True]  \\
                                                                                            & dropout\_list            & {[}-1, 0.5, -1, 0.5, -1]                \\
        \hline
    \end{tabular}
    }
\end{table}


The \emph{DynamicNet}-based \emph{EEGNet} model was trained as in its original paper~\cite{EEGNet_paper}: 500 epochs, ADAM optimizer (default parameters) and negative log likelihood loss (NLLLoss). A RTX 2070 GPU was used. 

\subsection{Cross-validation}
To investigate the performance of the \emph{EEGNet} over different contexts (intra- and cross-subject), we tested the following two different cross-validation schemes: 
\begin{itemize}
    \item \textit{Subject-dependent (intra-subject).} Here, one subject at a time was considered (with his/her data split into a training set and a test set).
    \item \textit{Cross-subject.} Here, two variants were implemented: (1) \emph{leave-one-subject-out} cross-validation, where the DL model was trained using data from all subjects except one, whose data were used as test set, and (2) \emph{mixed-up training}; in this case, first the data from each subject were split into a subject-related training set and subject-related test set; then, all subject-related training sets (from all subjects) were merged together in a mixed-up training set that was used to train the DL model; finally, each subject-related test set was used to test the DL model.
\end{itemize}


\section{Results and Discussion}
\label{sec:4_Results and discussions}
In this section, we report the results of our implementation of the \emph{EEGNet} through \emph{DynamicNet}, with the two above-mentioned different cross-validation schemes. Moreover, we emphasize how they can affect the model's performance and we compare our results with the state-of-the-art. All classification results are reported in Table~\ref{tab:EEGNet accuracy}.
As a baseline comparison, we also reported the results obtained using our implementation of the gold-standard FBCSP\footnote{Github repository FBCSP: \href{https://github.com/jesus-333/FBCSP-Python}{https://github.com/jesus-333/FBCSP-Python}}.

\subsection{Dataset}
\label{subsec:dataset}
We used the \emph{dataset 2a} from the BCI Competition IV~\cite{dataset_BCI_competition}. Particularly, the dataset includes 22-channel EEG recordings from 9 subjects performing the imagination of 4 different movements, i.e., right and left hand, feet and tongue movements. Therefore, the dataset is a 4-class MI EEG dataset. The sampling frequency is $\SI{250}{\hertz}$. For each subject, 288 training trials and 288 testing trials have been collected (a total of 5184 trials were available in the dataset).
The dataset has been preliminarily pre-processed by the authors, applying a band-pass filter in $0.5-100\SI{}{\hertz}$ and a notch filter around 50 Hz.
From each trial of each subject, $\SI{4}{\second}$ EEG segment is extracted.

\subsection{Classification accuracy}
%


\begin{table*}[h]
    \centering
    \caption{4-class classification accuracy and standard deviation for \emph{EEGNet} and FBCSP for different training strategies on raw data. The measures are reported on a scale from 0 to 100\%. * on the AVG row means that t-test between EEGNet and FBCSP corresponding values resulted in a p-value < 0.005.}
    \label{tab:EEGNet accuracy}
    \resizebox{\linewidth}{!}{%
    \begin{tabular}{|c|ccc||ccc|}  
    \cline{2-7}
    \multicolumn{1}{>{\centering\hspace{0pt}}m{0.056\linewidth}|}{} & EEGNet (Single) & EEGNet (mixed-up) & EEGNet (Cross) & FBCSP~
      (single) & FBCSP (mixed-up) & FBCSP (Cross)  \\ 
    \hline
    1                                                               & 74.76±6.43      & 81.42±1.53   & 67.22±2.87     & 69.97±1.13        & 57.05±1.17  & 41.75±1.11     \\
    2                                                               & 52.81±4.64      & 60.03±2.24   & 48.82±2.31     & 55.27±0.81        & 42.26±0.68  & 28.82±1.23     \\
    3                                                               & 84.58±3.22      & 89.34±2.24   & 73.4±1.11      & 79.28±1.73        & 53.44±1.81  & 47.74±2.03     \\
    4                                                               & 53.19±4.12      & 69.86±2.56   & 58.82±1.78     & 53.07±0.84        & 38.89±0.82  & 36.72±0.83     \\
    5                                                               & 69.86±2.81      & 60.58±14.54  & 42.78±7.24     & 46.35±0.7         & 31.35±0.71  & 25±0.55        \\
    6                                                               & 53.82±3.31      & 61.7±1.66    & 48.13±1.02     & 40.86±0.88        & 34.06±0.9   & 26.04±1.13     \\
    7                                                               & 74.51±6.63      & 80.73±2.51   & 68.75±3.31     & 76.85±0.66        & 37.22±0.73  & 27.17±0.67     \\
    8                                                               & 73.47±3.17      & 82.33±1.23   & 71.46±1.87     & 69.73±0.88        & 62.92±1.63  & 36.72±0.15     \\
    9                                                               & 73.92±6.89      & 66.49±2.61   & 54.79±2.28     & 71.7±1.41         & 55.49±1.57  & 36.11±1.7      \\ 
    \hline
    AVG         & 67.88±1.19      & 72.5±1.8*         & 59.35±1.97*  & 62.56±1.25        & 45.85±0.23  & 34.01±0.29     \\
    \hline
    \end{tabular}
    }
\end{table*}

The DL  model reached an average accuracy across all subjects of 68\% with the \textit{subject-dependent training} strategy, outperforming the FBCSP method that achieved a 63\% accuracy (see Table~\ref{tab:EEGNet accuracy}).
When the \textit{mixed-up training} strategy was used, the accuracy of the DL model increased to 73\%, while the accuracy obtained through FBCSP dropped to 46\%. The improvement provided by the DL model can be due to the increase in the amount of data used for the training. 
EEGNet managed to outperform FBCSP also with the \textit{cross-subject} training strategy (EEGNet achieved an accuracy of 59\% while the FBCSP reached 34\%).
%


\subsection{Cross-subject investigation}

Our cross-subject results are higher compared to the work of Lawhern \emph{et al.}~\cite{EEGNet_paper} (40\%) and also 4\% higher than~\cite{Amin_MCNN} (55.3\%).
These two works are the most similar to ours and therefore the easiest to compare with, since they are the only ones that used 4 classes.

To understand the difference between our results and the results of Lawhern \emph{et al.}, we conducted further tests using their training strategy (5 subjects for training, 3 for validation set and 1 for testing). The initial hypothesis was that the lower results were due to the smaller training set. Instead, we found out that the discriminant factor was the data preprocessing. We used their same training strategy with both raw data and preprocessed data. For both cases, we repeated the experiment 10 times. We obtained an average accuracy of 53\% for the raw data (in line with our previous results) and an average accuracy of 45\% with the preprocessed data (in line with~\cite{EEGNet_paper}). This suggests that MI deep learning models could be very sensitive to subject-to-subject variability in data contamination and amplitude range, and that the filtering and the normalization could remove subject-independent features while leaving only subject-dependent features. This hypothesis is also supported by the fact that, generally, filters and normalization either have no effect or increase the accuracy in intra-subject training.

Another interesting fact is the general increase in accuracy when we used the \textit{mixed-up} strategy, i.e., we used the training set of all subjects during the training. This approach could still be seen as a cross-subject training because the model is fed with data of multiple subjects and the test set includes data from a single one, only. This suggests that, as the dataset size increases, provided that also data from multiple subjects are included, EEGNet has the ability to extract more robust features, especially if compared with FBCSP. This increase in accuracy is also in line with the surveyed literature on DL, where the performance of the models increase with the size of the dataset. 

\section{Conclusions}
\label{sec:5_Conclusions}
In this paper we have surveyed several deep learning models for MI EEG data classification. Deep learning can achieve accuracy between 67\% and 76\% in 4-class intra-subject classification over the publicly available \emph{dataset 2a} (the most common dataset for MI classification). Deep learning was able to match and outperform FBCSP in intra-subject classification.
In binary cross-subject classification, deep learning models performed well with accuracy of 70\% (\emph{dataset 2a}), 76\% (\emph{dataset 2b}) and 83\% (\emph{eegmmidb}).
From our survey, we can conclude that room for improvement remains in multi-class cross-subject MI classification, where less studies are available, and the performance are still relatively low (\textasciitilde{}60\%).
Moreover, from our results, deep learning outperformed FBCSP in 4-class MI classification with an improvement of about 25\%, but further studies are needed to allow a wider use of this model in real-world scenarios. 
Finally, \emph{DynamicNet} showed to be an effective tool for the quick implementation of simple deep learning models, even though it is still in a preliminary version. 

\section*{Acknowledgment}
This work has been financially supported by the Project B-RELIABLE (PTDC/EEI-AUT/30935/2017), with FEDER/OE funding through programs CENTRO2020 and Portuguese Foundation for Science and Technology (FCT), and by ISR-UC  UIDB/00048/2020 through FCT funding. 
This work was also partially supported by REPAC project, funded by the University of Padova under the initiative SID-Networking 2019, and by MUR (Ministry of University and Research) under the initiative Departments of Excellence (Law 232/2016).

\bibliographystyle{IEEEtran}
\bibliography{biblio}

\begin{thebibliography}{10}
\providecommand{\url}[1]{#1}
\csname url@samestyle\endcsname
\providecommand{\newblock}{\relax}
\providecommand{\bibinfo}[2]{#2}
\providecommand{\BIBentrySTDinterwordspacing}{\spaceskip=0pt\relax}
\providecommand{\BIBentryALTinterwordstretchfactor}{4}
\providecommand{\BIBentryALTinterwordspacing}{\spaceskip=\fontdimen2\font plus
\BIBentryALTinterwordstretchfactor\fontdimen3\font minus
  \fontdimen4\font\relax}
\providecommand{\BIBforeignlanguage}[2]{{%
\expandafter\ifx\csname l@#1\endcsname\relax
\typeout{** WARNING: IEEEtran.bst: No hyphenation pattern has been}%
\typeout{** loaded for the language `#1'. Using the pattern for}%
\typeout{** the default language instead.}%
\else
\language=\csname l@#1\endcsname
\fi
#2}}
\providecommand{\BIBdecl}{\relax}
\BIBdecl

\bibitem{wolpaw_BCI}
\BIBentryALTinterwordspacing
J.~R. Wolpaw, D.~J. McFarland, G.~W. Neat, and C.~A. Forneris, ``An eeg-based
  brain-computer interface for cursor control,'' \emph{Electroencephalography
  and Clinical Neurophysiology}, vol.~78, no.~3, pp. 252--259, 1991. [Online].
  Available:
  \url{https://www.sciencedirect.com/science/article/pii/001346949190040B}
\BIBentrySTDinterwordspacing

\bibitem{Silvoni2013}
S.~Silvoni, M.~Cavinato, C.~Volpato, G.~Cisotto, C.~Genna, M.~Agostini,
  A.~Turolla, A.~Ramos-Murguialday, and F.~Piccione, ``Kinematic and
  neurophysiological consequences of an assisted-force-feedback brain-machine
  interface training: a case study,'' \emph{Frontiers in neurology}, vol.~4, p.
  173, 2013.

\bibitem{BCI_comunication}
\BIBentryALTinterwordspacing
M.~S. Hossain, S.~U. Amin, M.~Alsulaiman, and G.~Muhammad, ``Applying deep
  learning for epilepsy seizure detection and brain mapping visualization,''
  \emph{ACM Trans. Multimedia Comput. Commun. Appl.}, vol.~15, no.~1s, Feb.
  2019. [Online]. Available: \url{https://doi.org/10.1145/3241056}
\BIBentrySTDinterwordspacing

\bibitem{BCI_wheelchair}
L.~Tonin, T.~Carlson, R.~Leeb, and J.~del R.~Millán, ``Brain-controlled
  telepresence robot by motor-disabled people,'' in \emph{2011 Annual
  International Conference of the IEEE Engineering in Medicine and Biology
  Society}, 2011, pp. 4227--4230.

\bibitem{BCI_gaming}
A.~Nijholt, ``Bci for games: A `state of the art' survey,'' in
  \emph{Entertainment Computing - ICEC 2008}, S.~M. Stevens and S.~J.
  Saldamarco, Eds.\hskip 1em plus 0.5em minus 0.4em\relax Berlin, Heidelberg:
  Springer Berlin Heidelberg, 2009, pp. 225--228.

\bibitem{BCI_rehabilitation}
R.~Mane, T.~Chouhan, and C.~Guan, ``Bci for stroke rehabilitation: Motor and
  beyond,'' \emph{Journal of Neural Engineering}, vol.~17, 06 2020.

\bibitem{CSP_original_paper}
\BIBentryALTinterwordspacing
Z.~Koles, M.~Lazar, and S.~Zhou, ``Spatial patterns underlying population
  differences in the background eeg,'' \emph{Brain topography}, vol.~2, no.~4,
  p. 275—284, 1990. [Online]. Available:
  \url{https://doi.org/10.1007/bf01129656}
\BIBentrySTDinterwordspacing

\bibitem{FBCSP}
{Kai Keng Ang}, {Zheng Yang Chin}, {Haihong Zhang}, and {Cuntai Guan}, ``Filter
  bank common spatial pattern (fbcsp) in brain-computer interface,'' in
  \emph{2008 IEEE International Joint Conference on Neural Networks (IEEE World
  Congress on Computational Intelligence)}, 2008, pp. 2390--2397.

\bibitem{FBCSP_results_citation}
S.~M. Christensen, N.~S. Holm, and S.~Puthusserypady, ``An improved five class
  mi based bci scheme for drone control using filter bank csp,'' in \emph{2019
  7th International Winter Conference on Brain-Computer Interface (BCI)}, 2019,
  pp. 1--6.

\bibitem{EEGNet_paper}
V.~Lawhern, A.~Solon, N.~Waytowich, S.~Gordon, C.~Hung, and B.~Lance, ``Eegnet:
  A compact convolutional network for eeg-based brain-computer interfaces,''
  \emph{Journal of Neural Engineering}, vol.~15, 11 2016.

\bibitem{Schirrmeister_EEG_CNN}
\BIBentryALTinterwordspacing
R.~T. Schirrmeister, J.~T. Springenberg, L.~D.~J. Fiederer, M.~Glasstetter,
  K.~Eggensperger, M.~Tangermann, F.~Hutter, W.~Burgard, and T.~Ball, ``Deep
  learning with convolutional neural networks for eeg decoding and
  visualization,'' \emph{Human Brain Mapping}, aug 2017. [Online]. Available:
  \url{http://dx.doi.org/10.1002/hbm.23730}
\BIBentrySTDinterwordspacing

\bibitem{Sakhavi_EEG_CNN}
S.~{Sakhavi}, C.~{Guan}, and S.~{Yan}, ``Learning temporal information for
  brain-computer interface using convolutional neural networks,'' \emph{IEEE
  Transactions on Neural Networks and Learning Systems}, vol.~29, no.~11, pp.
  5619--5629, 2018.

\bibitem{Bressan2021}
G.~Bressan, G.~Cisotto, G.~R. M{\"u}ller-Putz, and S.~C. Wriessnegger, ``Deep
  learning-based classification of fine hand movements from low frequency
  eeg,'' \emph{Future Internet}, vol.~13, no.~5, p. 103, 2021.

\bibitem{Amin_MCNN}
\BIBentryALTinterwordspacing
S.~U. Amin, M.~Alsulaiman, G.~Muhammad, M.~A. Mekhtiche, and M.~{Shamim
  Hossain}, ``Deep learning for eeg motor imagery classification based on
  multi-layer cnns feature fusion,'' \emph{Future Generation Computer Systems},
  vol. 101, pp. 542--554, 2019. [Online]. Available:
  \url{https://www.sciencedirect.com/science/article/pii/S0167739X19306077}
\BIBentrySTDinterwordspacing

\bibitem{Borra_SincShallowNet}
\BIBentryALTinterwordspacing
D.~Borra, S.~Fantozzi, and E.~Magosso, ``Interpretable and lightweight
  convolutional neural network for eeg decoding: Application to movement
  execution and imagination,'' \emph{Neural Networks}, vol. 129, pp. 55--74,
  2020. [Online]. Available:
  \url{https://www.sciencedirect.com/science/article/pii/S0893608020302021}
\BIBentrySTDinterwordspacing

\bibitem{Xu_cross_dataset}
\BIBentryALTinterwordspacing
L.~Xu, M.~Xu, Y.~Ke, X.~An, S.~Liu, and D.~Ming, ``Cross-dataset variability
  problem in eeg decoding with deep learning,'' \emph{Frontiers in Human
  Neuroscience}, vol.~14, p. 103, 2020. [Online]. Available:
  \url{https://www.frontiersin.org/article/10.3389/fnhum.2020.00103}
\BIBentrySTDinterwordspacing

\bibitem{EEGNet_Fusion}
\BIBentryALTinterwordspacing
K.~Roots, Y.~Muhammad, and N.~Muhammad, ``Fusion convolutional neural network
  for cross-subject eeg motor imagery classification,'' \emph{Computers},
  vol.~9, no.~3, 2020. [Online]. Available:
  \url{https://www.mdpi.com/2073-431X/9/3/72}
\BIBentrySTDinterwordspacing

\bibitem{An_FewshotNetwork}
S.~An, S.~Kim, P.~Chikontwe, and S.~H. Park, ``Few-shot relation learning with
  attention for eeg-based motor imagery classification,'' 2020.

\bibitem{incpet_EEGNet_paper}
M.~Riyad, M.~Khalil, and A.~Adib, ``Incep-eegnet: A convnet for motor imagery
  decoding,'' in \emph{Image and Signal Processing}, A.~El~Moataz, D.~Mammass,
  A.~Mansouri, and F.~Nouboud, Eds.\hskip 1em plus 0.5em minus 0.4em\relax
  Cham: Springer International Publishing, 2020, pp. 103--111.

\bibitem{dataset_BCI_competition}
B.~Blankertz, G.~Dornhege, M.~Krauledat, K.-R. Müller, and G.~Curio, ``The
  non-invasive berlin brain-computer interface: Fast acquisition of effective
  performance in untrained subjects,'' \emph{NeuroImage}, vol.~37, pp. 539--50,
  09 2007.

\bibitem{eegmidb}
G.~Schalk, D.~McFarland, T.~Hinterberger, N.~Birbaumer, and J.~Wolpaw,
  ``Bci2000: a general-purpose brain-computer interface (bci) system,''
  \emph{IEEE Transactions on Biomedical Engineering}, vol.~51, no.~6, pp.
  1034--1043, 2004.

\bibitem{gradCAM_paper}
\BIBentryALTinterwordspacing
R.~R. Selvaraju, M.~Cogswell, A.~Das, R.~Vedantam, D.~Parikh, and D.~Batra,
  ``Grad-cam: Visual explanations from deep networks via gradient-based
  localization,'' \emph{International Journal of Computer Vision}, vol. 128,
  no.~2, p. 336–359, Oct 2019. [Online]. Available:
  \url{http://dx.doi.org/10.1007/s11263-019-01228-7}
\BIBentrySTDinterwordspacing

\bibitem{SHAP_paper}
\BIBentryALTinterwordspacing
S.~M. Lundberg and S.-I. Lee, ``A unified approach to interpreting model
  predictions,'' pp. 4765--4774, 2017. [Online]. Available:
  \url{https://proceedings.neurips.cc/paper/2017/file/8a20a8621978632d76c43dfd28b67767-Paper.pdf}
\BIBentrySTDinterwordspacing

\end{thebibliography}

\end{document}